\documentclass[10pt,twocolumn,letterpaper]{article}

\usepackage{cvpr}              

\usepackage[dvipsnames]{xcolor}
\usepackage{nicematrix,tikz}
\usepackage{algorithm}
\usepackage{algpseudocode}
\usepackage{graphicx}
\usepackage{bbm}
\usepackage{multirow}

\usepackage{colortbl}
\usepackage{booktabs}
\NiceMatrixOptions
  {
    custom-line = 
     {
       letter = : ,
       command = dashedline , 
       ccommand = cdashedline ,
       tikz = dashed
     }
  }

\definecolor{cvprblue}{rgb}{0.21,0.49,0.74}
\usepackage[pagebackref,breaklinks,colorlinks,allcolors=cvprblue]{hyperref}
\usepackage{arydshln}
\def\paperID{9025} 
\def\confName{CVPR}
\def\confYear{2025}

\title{Universal Domain Adaptation for Semantic Segmentation}

\author{Seun-An Choe\textsuperscript{1}
\and
Keon-Hee Park\textsuperscript{1}
\and
Jinwoo Choi\textsuperscript{1}
\and
Gyeong-Moon Park\textsuperscript{2}\thanks{Corresponding author.} \\ \\
\textsuperscript{1}Kyung Hee University, Yongin, Republic of Korea \\
\textsuperscript{2}Korea University, Seoul, Republic of Korea\\
{\tt\small \textsuperscript{1}\{dragoon0905, pgh2874, jinwoochoi\}@khu.ac.kr} \quad {\tt\small \textsuperscript{2}gm-park@korea.ac.kr}
}

\makeatletter
\renewcommand*{\@fnsymbol}[1]{\ensuremath{\ifcase#1\or \dagger\or \ddagger\or
   \mathsection\or \mathparagraph\or \|\or **\or \dagger\dagger
   \or \ddagger\ddagger \else\@ctrerr\fi}}
\makeatother

\begin{document}
\maketitle
\begin{abstract}
Unsupervised domain adaptation for semantic segmentation (UDA-SS) aims to transfer knowledge from labeled source data to unlabeled target data. However, traditional UDA-SS methods assume that category settings between source and target domains are known, which is unrealistic in real-world scenarios. This leads to performance degradation if private classes exist.
To address this limitation, we propose Universal Domain Adaptation for Semantic Segmentation (UniDA-SS), achieving robust adaptation even without prior knowledge of category settings. We define the problem in the UniDA-SS scenario as low confidence scores of common classes in the target domain, which leads to confusion with private classes. To solve this problem, we propose UniMAP: \textbf{Uni}DA-SS with Image \textbf{Ma}tching and \textbf{P}rototype-based Distinction, a novel framework composed of two key components.
First, Domain-Specific Prototype-based Distinction (DSPD) divides each class into two domain-specific prototypes, enabling finer separation of domain-specific features and enhancing the identification of common classes across domains. Second, Target-based Image Matching (TIM) selects a source image containing the most common-class pixels based on the target pseudo-label and pairs it in a batch to promote effective learning of common classes. We also introduce a new UniDA-SS benchmark and demonstrate through various experiments that UniMAP significantly outperforms baselines. The code is available at \url{https://github.com/KU-VGI/UniMAP}.

\end{abstract}    
\section{Introduction}
\label{sec:intro}

\begin{figure}[t] 

\centering
\includegraphics[width=1\columnwidth]{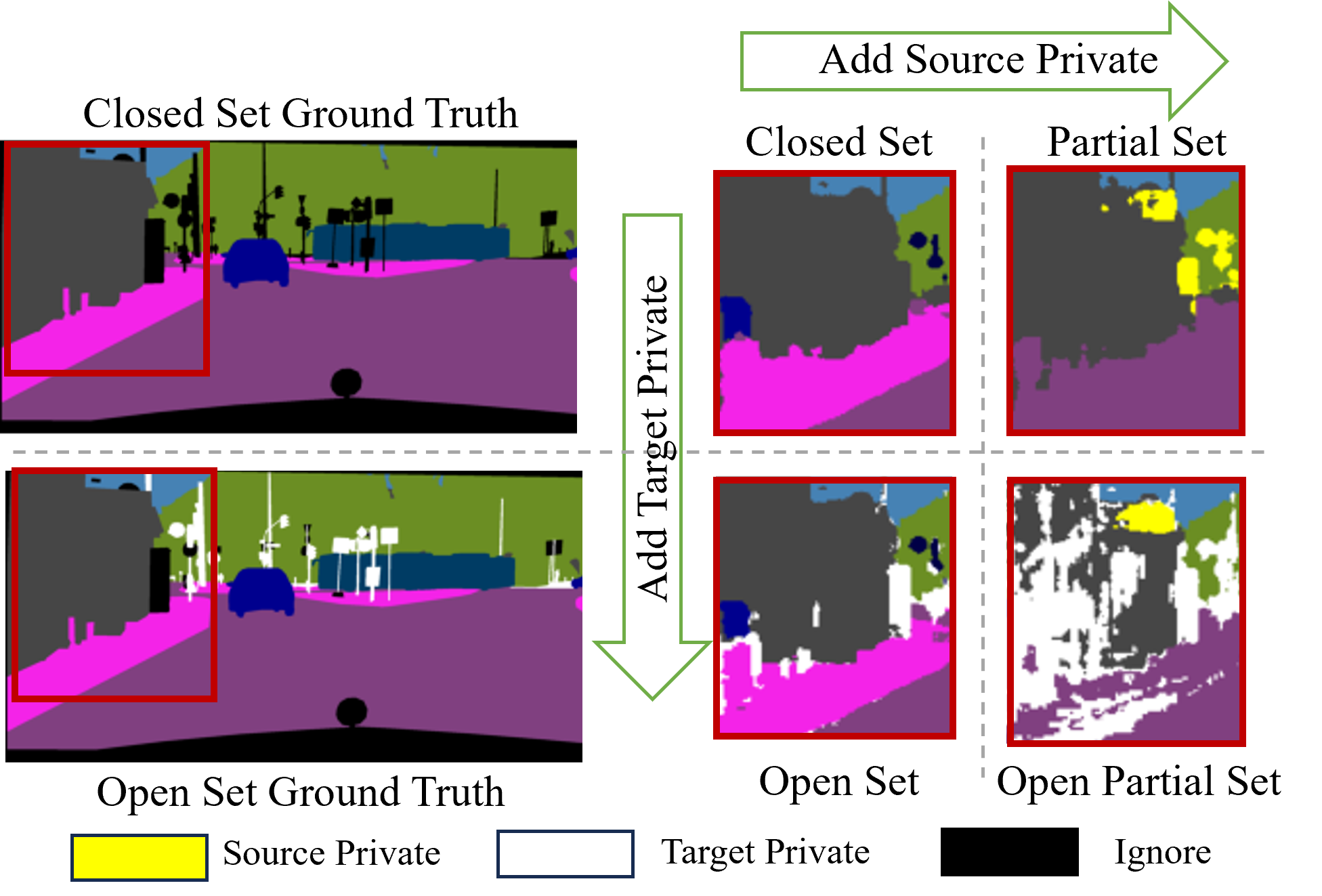}
\vspace{-6mm}

\caption{Visualization results of the UDA-SS models across different scenarios. We select MIC and BUS, which achieve the best performance in CDA-SS and ODA-SS, respectively, and visualize their results in PDA-SS and OPDA-SS. The images illustrate the performance degradation caused by the introduction of source-private classes. \vspace{-7mm}}
\label{fig:motive}

\end{figure}
Semantic segmentation is a fundamental computer vision task that predicts the class of each pixel in an image and is essential in fields like autonomous driving, medical imaging, and human-machine interaction. However, training segmentation models requires pixel-level annotations, which are costly and time-consuming. To address this, researchers have explored Unsupervised Domain Adaptation for Semantic Segmentation (UDA-SS) methods, which aim to learn domain-invariant representations from labeled synthetic data (source) to unlabeled real-world data (target).

While UDA-SS has shown effectiveness in addressing domain shift, existing UDA-SS methods rely on the assumption that source and target categories are known in advance. This assumption is often impractical in real-world scenarios, as target labels are typically unavailable. As a result, the target domain frequently includes unseen classes that are absent in the source domain (target-private classes), or conversely, the source domain may contain classes not found in the target domain (source-private classes). This limitation can lead to negative transfer, where models incorrectly align source-private classes with the target domain, resulting in significant performance degradation. To address these challenges, we introduce a new Universal Domain Adaptation for Semantic Segmentation (UniDA-SS) scenario, enabling adaptation without prior knowledge of category configurations and classifying target samples as ``unknown" if they contain target-private classes.

To understand the challenges posed by the UniDA-SS scenario, we first evaluate the performance of existing UDA-SS methods under various domain adaptation settings. Figure ~\ref{fig:motive} presents qualitative results of UDA-SS methods across various scenarios. Specifically, we select MIC~\cite{hoyer2023mic} and BUS~\cite{choe2024open} as representative models for Closed Set Domain Adaptation (CDA-SS) and Open Set Domain Adaptation (ODA-SS), respectively, and analyze their performance in Partial Domain Adaptation (PDA-SS) and Open Partial Domain Adaptation (OPDA-SS) settings. CDA-SS assumes that the source and target domains share the same set of classes, while ODA-SS contains target-private classes that do not exist in the source domain. PDA-SS, on the other hand, assumes that the target domain contains only a subset of the source classes. OPDA-SS extends PDA-SS by adopting the open-set characteristic of ODA-SS, where both source-private and target-private classes exist simultaneously.

These evaluations reveal that adding source-private classes in transitions from CDA to PDA and from ODA to OPDA degrades performance. In PDA, common classes like “buildings” are often misclassified as source-private, while ``sidewalk" regions are mistakenly predicted as ``road". Similarly, in OPDA, target-private regions are frequently confused with source-private or common classes. 
Most state-of-the-art UDA-SS methods depend on self-training with target pseudo-labels, heavily relying on pseudo-label confidence scores. Particularly, in ODA-SS scenarios such as BUS, confidence scores are also used to assign unknown pseudo-labels. When source-private classes are present, their feature similarity to some common classes increases, leading to a reduction in pseudo-label confidence. As a result, common classes may not be effectively learned, and if the confidence score drops below a certain threshold ($\tau_p$), common classes are often misassigned as target-private classes. This misassignment hinders the effective learning of both common and target-private classes, further degrading adaptation performance.

To mitigate this issue,  we propose a novel framework, UniMAP, \textbf{Uni}DA-SS with Image \textbf{Ma}tching and \textbf{P}rototype-based Distinction, aim to increase the confidence scores of common classes under unknown category settings. First, we introduce a Domain-Specific Prototype-based Distinction (DSPD) to distinguish between common classes and source-private classes while considering variations of common classes between the source and target domains. Unlike conventional UDA-SS methods, which treat common classes as identical across domains, DSPD assigns two prototypes per class—one for source and one for target—to learn with one class while capturing domain-specific features. This approach enables independent learning of source and target-specific features, enhancing confidence scores for target predictions. Additionally, since common class pixel embeddings will have similar relative distances to the source and target prototypes, and the private class will be relatively close to any one prototype, we can use this to distinguish between common and private classes and assign higher weights to pixel embeddings that are more likely to belong to the common classes.

Second, to increase the confidence scores of the common classes, it is crucial to increase their pixel presence during training for robust domain-invariant representation. However, source-private classes often reduce the focus on common classes, hindering effective adaptation. To address this issue, we propose Target-based Image Matching (TIM), which prioritizes source images with the highest number of common class pixels based on target pseudo-labels. TIM compares target pseudo-labels and source ground truth at the pixel level, selecting the source images that overlap the most in common classes to pair with the target image in a single batch. This matching strategy facilitates domain-invariant learning of common classes, improving performance in a variety of scenarios. We also utilize a class-wise weighting strategy based on the target class distribution to assign higher weights to rare classes to address the class imbalance problem.

We summarize our main contributions as follows:
\begin{itemize}
\item We introduce a new task the Universal Domain Adaptation for Semantic Segmentation (UniDA-SS) task for the first time. To address this, we propose a novel framework called UniMAP, short for \textbf{Uni}DA-SS with Image \textbf{Ma}tching and \textbf{P}rototype-based Distinction.

\item To enhance pseudo-label confidence in the target domain, we propose Domain-Specific Prototype-based Distinction (DSPD), a pixel-level clustering approach that utilizes domain-specific prototypes to distinguish between common and private classes.

\item We propose Target-based Image Matching (TIM), which enhances domain-invariant learning by prioritizing source images rich in common class pixels.

\item We demonstrate the superiority of our approach by achieving state-of-the-art performance compared to existing UDA-SS methods through extensive experiments.
\end{itemize}
\section{Related Work}
\subsection{Semantic Segmentation.}
Semantic segmentation aims to classify each pixel in an image into a specific semantic. A foundational approach, Fully Convolutional Networks (FCNs)~\cite{long2015fully}, has demonstrated impressive performance in this task. To enhance contextual understanding, subsequent works have introduced methods such as dilated convolutions ~\cite{chen2017deeplab}, global pooling ~\cite{liu2015parsenet}, pyramid pooling ~\cite{zhao2017pyramid}, and attention mechanisms~\cite{zhao2018psanet, zhu2019asymmetric}.  More recently, transformer-based methods have achieved significant performance gains ~\cite{xie2021segformer, zheng2021rethinking}.  Despite various studies, semantic segmentation models are still vulnerable to domain shifts or category shifts. To address this issue, we propose a universal domain adaptation for semantic segmentation that handles domain shifts and category shifts.

\subsection{Unsupervised Domain Adaptation for Semantic Segmentation.}
Unsupervised Domain Adaptation (UDA) aims to leverage labeled source data to achieve high performance on unlabeled target data. Existing UDA methods for semantic segmentation can be categorized into two approaches: adversarial learning-based and self-training. Adversarial learning-based methods~\cite{tsai2018learning,hong2018conditional,kim2020learning,pan2020unsupervised,tsai2019domain,chen2019synergistic,du2019ssf} use an adversarial domain classifier to learn domain-invariant features. Self-training methods~\cite{melas2021pixmatch,hoyer2022daformer,hoyer2022hrda,zou2018unsupervised,chen2019domain,zou2019confidence, wang2021domain,lian2019constructing,li2019bidirectional,wang2021uncertainty,zhang2021prototypical, tranheden2021dacs} assign pseudo-labels to each pixel in the target domain using confidence thresholding, and several self-training approaches further enhance target domain performance by re-training the model with these pseudo-labels. Although UDA allows the model to be trained on the target domain without annotations, it requires prior knowledge of class overlap between the source and target domains, which limits the model's applicability and generalizability. To overcome this limitation, we propose a universal domain adaptation approach for semantic segmentation, where the model can adapt to the target domain without requiring prior knowledge of class overlap.

\subsection{Universal Domain Adaptation in Classification}
Universal Domain Adaptation (UniDA)~\cite{you2019universal} was introduced to address various domain adaptation settings, such as closed-set, open-set, and partial domain adaptation. UniDA is a more challenging scenario because it operates without prior knowledge of the category configuration of the source and target domains. To tackle UniDA in classification tasks, prior works have focused on computing confidence scores for known classes and treating samples with lower scores as unknowns. CMU~\cite{fu2020learning} proposed a thresholding function, while ROS~\cite{bucci2020effectiveness} used the mean confidence score as a threshold, which results in neglecting about half of the target data as unknowns. DANCE~\cite{saito2020universal} set a threshold based on the number of classes in the source domain. OVANet~\cite{saito2021ovanet} introduced training a threshold using source samples and adapting it to the target domain. While UniDA has been extensively studied in the context of classification tasks, it remains underexplored in semantic segmentation, which requires a higher level of visual understanding due to the need for pixel-wise classification. In this work, we aim to investigate UniDA for semantic segmentation.



\section{Method}
\label{sec:Method}
\begin{figure*}[t] 

\centering
\includegraphics[width=1\textwidth]{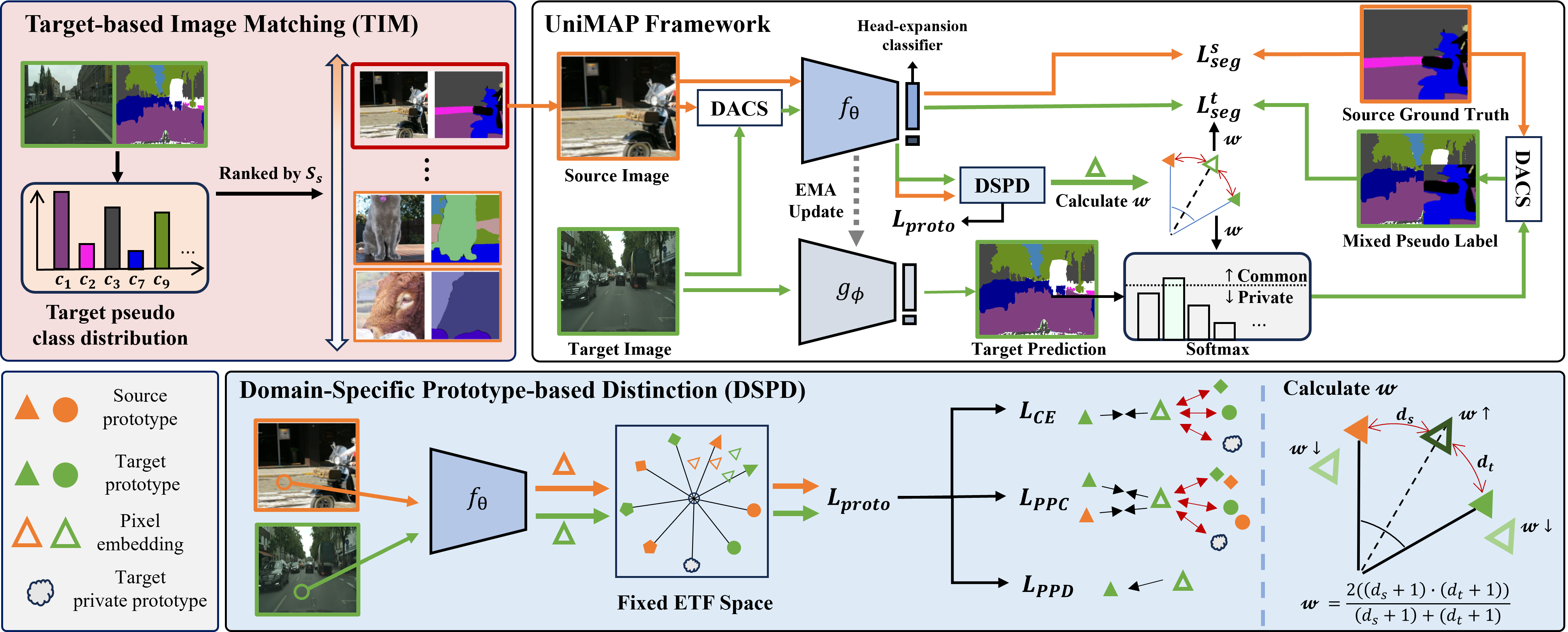}
\vspace{-6mm}

\caption{Overview of our proposed method, UniMAP. The top right illustrates the main training framework. The model is optimized with three main losses: the supervised segmentation loss on the source domain $L^s_{seg}$, the pseudo-label guided loss on the target domain $L^t_{seg}$ using DACS~\cite{tranheden2021dacs}, which is a domain mixing technique, and $L_{proto}$, the prototype-based loss $L_{proto}$ computed in a fixed ETF space~\cite{papyan2020prevalence}. $L_{proto}$ consists of three losses, which allows the prototype to have domain-specific information. Pixel-wise weight scaling factor $w$, is derived based on the relative distance between source and target prototypes, assigning higher weights to common classes that align well with both prototypes. These weights are used in generating target pseudo-labels and the target loss $L_{seg}^t$. On the top left is the framework of TIM. It computes the class distribution of the target pseudo-label and ranks source images based on class overlap using the similarity score $S_{s}$. The top-ranked source image is selected and paired with the target image in each training batch. \vspace{-3mm}}
\vspace{-3mm}
\label{fig:main} 

\end{figure*}
\subsection{Problem Formulation}
In the UniDA-SS scenario, the goal is to transfer knowledge from a labeled source domain \(D_s\)$=\{X_s, Y_s\}$ to an unlabeled target domain \(D_t\)$=\{X_t\}$. The model is trained on the source images \(X_s\)$=\{x_s^1, x_s^2, ..., x_s^{i_s}\}$ with the corresponding labels \(Y_s\)$=\{y_s^1, y_s^2, ..., y_s^{i_s}\}$ and the target images \(X_t\)$=\{x_t^1, x_t^2, ..., x_t^{i_t}\}$, where ground-truth labels are unavailable.
Each image \(x_s^{i_s} \in \mathbb{R}^{3 \times H \times W}\) and \(y_s^{i_s} \in \mathbb{R}^{C \times H \times W}\) represent an \({i_s}\)-th RGB image and its pixel-wise label. 
\(H\) and \(W\) are the height and width of the image, and \(C_s\) and \(C_t\) denote the sets of classes in the source and target domains, respectively. We aim to adapt the model to perform well on \(D_t\), even though there is no prior knowledge of class overlap between \(C_s\) and \(C_t\) given. We define \(C_c = C_s \cap C_t\) as the set of common classes, while  \(C_s\ \backslash C_c\) and \(C_t\ \backslash C_c\) represent the classes private to each domain, respectively. To handle target-private samples in \(C_t\ \backslash C_s\), we classify them as ``unknown" without prior knowledge of their identities. Under this formulation, UniDA-SS requires addressing two challenges: (1) to classify common classes in \(C_c\) correctly and (2) to detect target-private classes in \(C_t\ \backslash C_s\).

\subsection{Baseline}
We construct our UniDA-SS baseline by adopting a standard open-set self-training approach, partially following the ODA-SS formulation introduced in BUS~\cite{choe2024open}. BUS handles unknown target classes by appending an additional classification head node to predict an unknown class. In our baseline, we adopt the same structural design as BUS but remove refinement components and the use of attached private class masks, resulting in a setup suitable for UniDA-SS.

In this baseline, the number of classifier heads is set to $(C_s+1)$, where the $(C_s+1)$-th head corresponds to the unknown class. The segmentation network \(f_\theta\) is trained with the labeled source data using the following categorical cross-entropy loss $\mathcal{L}_{seg}^{s}$:
\begin{align}
\mathcal{L}_{seg}^{s} = -\sum_{j=1}^{H \cdot W}\sum_{c=1}^{C_s+1} {y}_{s}^{(j,c)} \log f_\theta(x_{s})^{(j,c)},
\label{eq:source_cleanloss}
\end{align}
where $j \in \{1, 2, ..., H \cdot W\}$ denotes the pixel index and $c \in \{1, 2, ..., C_s+1\}$ denotes the class index. The baseline utilizes a teacher network \(g_\phi\) to generate the target pseudo-labels. \(g_\phi\) is updated from \(f_\theta\) via exponential moving average (EMA)~\cite{tarvainen2017mean} with a smoothing factor \(\alpha\). The pseudo-label $\hat{y}_{tp}^{(j)}$ for the $j$-th pixel considering unknown assigned as follows:
\begin{align}
\hat{y}_{tp}^{(j)} = 
\begin{cases} 
c', & \text{if } \left( \max_{c'}g_\phi(x_t)^{(j,c')} \geq \tau_p \right) \\
C_s+1, & \text{otherwise}
\end{cases}
,
\end{align}
where $c^{'} \in \{1, 2, ..., C_s\}$ denotes a known classes and $\tau_p$ is a fixed threshold for assign unknown pseudo-labels. Then, we calculate the image-level reliability of the pseudo-label \(q_t\) as follows~\cite{tranheden2021dacs}:
\begin{align}
q_t = \frac{1}{H \cdot W} \sum_{j=1}^{H \cdot W} \left[ \max_{c'} g_\phi(x_t)^{(j, c')} \geq \tau_t \right],
\end{align}
where \(\tau_t\) is a hyperparameter. The network \(f_\theta\) is trained using the pseudo-labels and the corresponding confidence estimates with the using the weighted cross-entropy loss $\mathcal{L}_{seg}^{t}$:
\begin{align}
\mathcal{L}_{seg}^{t} = -\sum_{j=1}^{H \cdot W}\sum_{c=1}^{C_s+1} {q_t \cdot \hat{y}_{tp}^{(j,c)} \log f_\theta(x_{t})^{(j,c)}}.
\label{eq:target_loss}
\end{align}
Based on this baseline, we propose a novel framework called UniMAP, short for \textbf{Uni}DA-SS with Image \textbf{Ma}tching and \textbf{P}rototype-based Distinction.
\subsection{Domain-Specific Prototype-based Distinction}
\paragraph{Prototype-based Learning.} In conventional self-training-based UDA-SS methods, common classes from both the source and target domains are typically treated as a unified class, assuming identical feature representations. However, in practice, common classes often exhibit domain-specific features (e.g., road appearance and texture differences between Europe and India). To address this issue, we leverage the concept from ProtoSeg~\cite{zhou2022rethinking}. ProtoSeg uses multiple non-learnable prototypes per class to represent diverse features within the pixel embedding space, adequately capturing inter-class variance. Building on this idea, we assign two prototypes per class, one for the source and one for the target. This allows the model to capture domain-specific features for each class while still learning them as a unified class, effectively enhancing the confidence scores for common classes in the target domain. To ensure that the source and target prototypes maintain a stable distance, we use a fixed Simplex Equiangular Tight Frame (ETF)~\cite{papyan2020prevalence}, which guarantees equal cosine similarity and L2-norm across all prototype pairs. This structure enables consistent separation between the source and target prototypes, facilitating the learning of domain-specific features. The prototypes are defined as follows:
\begin{align}
{\{p_k}\}_{k=1}^{2C+1}=\sqrt{\frac{2C+1}{2C}}U(I_{2C+1}-\frac{1}{2C+1}1_{[2C+1]}1_{[2C+1]}^{\intercal}),
\end{align}
Each class has a pair of prototypes $p^c \in \{p^{c}_{s}, p^{c}_{t}\}$, with an additional prototype is defined for unknown classes $p^{C+1} \in \{p^{C+1}_t\}$. We employ three prototype-based loss functions adapted from ProtoSeg for each domain $D \in \{s,t\}$.
First, the cross entropy loss $\mathcal{L}_{CE}$ that moves the target closer to the corresponding prototype and further away from the rest of the prototypes as follows:
\begin{align}
\mathcal{L}_{CE}^{D} = -log \frac{exp(i^\intercal p_{D}^c)}{exp(i^\intercal p_{D}^c)+\sum_{c'\neq c}exp(i^{\intercal}p_{D}^{c'})},
\end{align}
where $i$ represents the L2-normalized pixel embedding, using the label for source pixels and the pseudo-label for target pixels to determine the corresponding class $c$.
Second, pixel-prototype contrastive learning strategy $\mathcal{L}_{PPC}$, which makes it closer to the corresponding prototype in the entire space and farther away from the rest as follows:
\begin{align}
\mathcal{L}_{PPC}^{D} = -log \frac{\sum_{p \in p^c} exp(i^\intercal p^c/\tau)}{\sum_{p \in p^c} exp(i^\intercal p^c/\tau)+\sum_{p^{-} \in P^{-}}exp(i^{\intercal}p^{-}/\tau)},
\end{align}
where $P^{-}$ denotes set of prototypes excluding $p^c$. Finally, Pixel-Prototype Distance Optimization $\mathcal{L}_{PPD}$ makes the distance of pixel embedding and prototype closer as:
\begin{align}
\mathcal{L}_{PPD}^{D} = (1-i^{\intercal} p_{D}^c )^2.
\end{align}
Therefore, we can organize $\mathcal{L}_{proto}$ as follows: 
\begin{align}
\mathcal{L}_{proto} = \mathcal{L}_{CE} + \lambda_{1} \mathcal{L}_{PPC} + \lambda_{2} \mathcal{L}_{PPD},
\end{align}
where $ \lambda_{1}$ and $\lambda_{2}$ denote hyperparameters.
Through the $\mathcal{L}_{proto}$, the model can capture domain-specific features while learning each class as a unified representation.
\paragraph{Prototype-based Weight Scaling.} We further utilize prototypes to distinguish between common class and source-private. As training progresses, common-class pixel embeddings tend to align with both source and target prototypes, whereas private-class embeddings align with only one. Thus, when an embedding is similarly close to both prototypes, it is likely to be from a common class. Based on this, we assign a pixel-wise weight scaling factor $w$ to reflect the likelihood of a pixel belonging to a common class:

\begin{align}
w = \frac{2(d_s+1)(d_t+1)}{(d_s+1)+(d_t+1)},
\end{align}
where $d_s$, $d_t$ denote cosine similarity between pixel embedding $i$ and the source and target prototypes $p^c_s, p^c_t$, respectively. The scaling factor $w$  is then applied to ~\Cref{eq:target_loss} during pseudo-label generation as follows:
\begin{align}
\mathcal{L}_{seg}^{t} = -\sum_{j=1}^{H \cdot W}\sum_{c=1}^{C+1} {w \cdot {q_t} \hat{y}_{tp}^{(j,c)} \log f_\theta(x_{t})^{(j,c)}},
\label{eq:target_loss_withw}
\end{align}

\begin{align}
\hat{y}_{tp}^{(j)} = 
\begin{cases} 
c', & \text{if } \left( \max_{c'}g_\phi(x_t)^{(j,c')} \cdot w \geq \tau_p \right) \\
C+1, & \text{otherwise}
\end{cases}
.
\end{align}
The above method mitigates the assignment of a common class to target-private in the target pseudo-label and enhances the learning of pixels with a high probability of a common class.

\subsection{Target-based Image Matching}

To increase the confidence score of common classes, it is important to include as many common classes as possible in the training to learn domain-invariant representation. However, when source-private classes are added, the proportion of learning common classes decreases, making it difficult to learn domain-invariant representation.
To solve this problem, we propose the Target-based Image Matching (TIM) method, which selects images containing as many common classes as possible from source images based on the classes appearing in the target pseudo-label. 
First, we calculate the proportion of each class present in the target pseudo-label $\hat{y}_{tp}$ as follows:
\begin{align}
f_c = \frac{n_c}{\sum_k n_k},
\end{align}
where $n_c$ denotes the number of pixels of class $c$ in $\hat{y}_{tp}$. Utilizing $f_c$ we calculate $\hat{f_c}$, which has a higher value for rare classes, as follows:
\begin{align}
\hat{f_c} = softmax(\frac{1-f_c}{T}),
\end{align}
where $T$ denotes temperature. For each source image through $\hat{f_c}$, we measure $S_s$ as follows:
\begin{align}
{S_s} = \sum_{c \in {c^*}}n_{c}^s\hat{f_{c}},
\end{align}
where $n^s_c$ denotes the number of pixels of class $c$ in $y_s$ and $c^*$ denotes set of overlapping classes between $y_s$ and $\hat{y}_{tp}$.
So, we select the source image with the highest ${S_s}$  and pair it with the corresponding target image in a training batch. This approach allows us to effectively learn domain-invariant representations for common classes, which can improve performance in a variety of scenarios. It also mitigates class imbalance by prioritizing source images that contain more pixels from rare common classes, guided by class weighting based on the target class distribution.

\begin{table*}[t]
\centering
\resizebox{\textwidth}{!}{%
\setlength{\tabcolsep}{6.pt}
\renewcommand{\arraystretch}{1.2}
\begin{tabular}{lcccccccccccc>{\columncolor[gray]{0.9}}c>{\columncolor[gray]{0.9}}c>{\columncolor[gray]{0.9}}c}
\specialrule{1.5pt}{0pt}{0pt}
\multicolumn{16}{c}{Pascal-Context $\rightarrow$ Cityscapes} \\ \hline
\multicolumn{1}{l|}{Method}                     & Road  & S.walk & Build. & Wall & Fence & Veget. & Sky   & Car   & \multicolumn{1}{l}{Truck} & Bus   & M.bike & \multicolumn{1}{c|}{Bike}  & Common & Private & H-score \\ \hline

\multicolumn{1}{l|}{UAN ~\cite{you2019universal}}      & 61.78 & 13.14  & 78.14  & 0.03 & 5.60  & 20.01  & 81.50 & 33.2  & 36.24 & 4.90  & 15.48  & \multicolumn{1}{c|}{13.01} & 31.93  & 4.30    & 7.47    \\
\multicolumn{1}{l|}{UniOT ~\cite{chang2022unified}}    & 62.34 & 15.64  & 75.69  & 0.05 & 4.61  & 21.50  & 78.10 & 34.3  & 35.04  & 5.94  & 12.98  & \multicolumn{1}{c|}{15.85} & 32.84  & 6.85    & 10.76   \\
\multicolumn{1}{l|}{MLNet ~\cite{lu2024mlnet}}    & 71.28 & 12.94  & 68.63  & 0.00 & 6.15  & 19.73  & 81.7  & 22.8  & 27.04   & 4.45  & 11.68  & \multicolumn{1}{c|}{10.72} & 30.81  & 6.43    & 10.61   \\ \hdashline
\multicolumn{1}{l|}{DAFormer ~\cite{hoyer2022daformer}} & 25.29 & 0.00   & 83.44  & 0.09 & 7.69  & 86.94  & \textbf{91.68} & \textbf{91.59} & \textbf{81.80}     & \textbf{66.18} & \textbf{55.66}  & \multicolumn{1}{c|}{60.49} & 54.24  & 4.43    & 8.20    \\
\multicolumn{1}{l|}{HRDA ~\cite{hoyer2022hrda}}     & 62.33 & 0.00  & 77.75  & \textbf{0.64} & 30.87 & 80.49  & 83.24 & 88.79 & 70.11     & 58.66 & 9.11   & \multicolumn{1}{c|}{21.75} & 51.89  & 8.55    & 14.68   \\
\multicolumn{1}{l|}{MIC ~\cite{hoyer2023mic}}      & 40.49 & 0.21   & 79.40  & 0.00 & 8.35  & 85.74  & 89.58 & 84.78 & 46.87     & 47.23 & 47.78  & \multicolumn{1}{c|}{53.59} & 48.67  & 7.85    & 13.51   \\ \hdashline
\multicolumn{1}{l|}{BUS ~\cite{choe2024open}}      & 77.90 & 0.01   & 85.26  & 0.00 & 31.16 & 87.12  & 88.43 & 89.94 & 64.51        & 53.71 & 50.22  & \multicolumn{1}{c|}{63.40} & 57.64  & 20.38   & 30.11   \\ \hline
\multicolumn{1}{l|}{UniMAP \textbf{(Ours)}}     & \textbf{84.15} & \textbf{16.77}  & \textbf{86.38}  & 0.00 & \textbf{35.12} & \textbf{88.26}  & 89.45 & 90.75 & 64.54       & 59.25 & 49.98  & \multicolumn{1}{c|}{\textbf{66.63}} & \textbf{60.94}  & \textbf{31.27}   & \textbf{41.33}   \\
\specialrule{1.5pt}{0pt}{0pt}
\end{tabular}%
}
\vspace{-3mm}
\caption{\label{tab:table-pascal_city} Semantic segmentation performance on Pascal-Context $\rightarrow$ Cityscapes OPDA-SS benchmarks. Our method outperformed baselines in common, private, and overall performance. White columns show individual common class scores, while ``Common'' in gray columns represents the average performance of common classes. The best results are highlighted in bold. \vspace{-3mm}}
\end{table*}

\begin{table*}[t]
\centering
\resizebox{\textwidth}{!}{%
\setlength{\tabcolsep}{2.pt}
\renewcommand{\arraystretch}{1.2}
\begin{tabular}{lccccccccccccccccc>{\columncolor[gray]{0.9}}c>{\columncolor[gray]{0.9}}c>{\columncolor[gray]{0.9}}c}
\specialrule{1.5pt}{0pt}{0pt}
\hline
\multicolumn{21}{c}{GTA5 $\rightarrow$ IDD} \\ \hline
\multicolumn{1}{l|}{Method}   & Road  & S.walk & Build. & Wall  & Fence & Pole  & Light & Sign  & Veget. & Sky   & Person & Rider & Car   & Truck & Bus   & M.bike & \multicolumn{1}{c|}{Bike}  & Common & Private & H-score \\ \hline

\multicolumn{1}{l|}{UAN ~\cite{you2019universal}}      & 97.38 & 61.33  & 62.24  & 36.27 & 16.41 & 24.11 & 8.96  & 58.29 & 78.82  & 94.15 & 57.06  & 30.09 & 68.98 & 72.92 & 42.66 & 64.93  & \multicolumn{1}{c|}{7.85}  & 49.20  & 3.14    & 5.92    \\
\multicolumn{1}{l|}{UniOT ~\cite{chang2022unified}}    &    96.99   &   41.19     &    63.61    &  34.63     &   18.96    &  28.35     & 3.96      &   54.07    &    72.9    &  92.89     &   53.9     &   32.36    &   81.82    &   72.85    &  63.84     &   63.28     & \multicolumn{1}{c|}{5.18}      &   51.82     &    7.44     &   13.01      \\
\multicolumn{1}{l|}{MLNet ~\cite{lu2024mlnet}}    &   95.59    &   9.87     &   55.53     &   17.26    &   12.14    &   12.69    &   5.81    &  64.13     &   72.69     &   91.57    &  0.00      &   17.92    &   69.59    &   65.65    &   50.35    &   60.76     & \multicolumn{1}{c|}{5.39}      & 41.58       &   4.23      &   7.68      \\ \hdashline
\multicolumn{1}{l|}{DAFormer ~\cite{hoyer2022daformer}} & 97.89 & 54.84  & 70.28  & 43.71 & 25.56 & 37.74 & 14.57 & 66.80 & 79.14  & 91.92 & 58.31  & 52.31 & 83.36 & 80.14 & 77.16 & 64.70  & \multicolumn{1}{c|}{21.54} & 52.05  & 21.07   & 29.99   \\
\multicolumn{1}{l|}{HRDA ~\cite{hoyer2022hrda}}     & 97.90 & 52.22  & 69.80  & 42.73 & 25.15 & 38.79 & 21.43 & 66.80 & 80.06  & 91.38 & 57.60  & 50.83 & 83.27 & 80.05 & 76.35 & 64.05  & \multicolumn{1}{c|}{20.07} & 57.83  & 22.47   & 32.69   \\
\multicolumn{1}{l|}{MIC ~\cite{hoyer2023mic}}      & 95.18 & 39.64  & 67.66  & 43.19 & 23.08 & 36.32 & 17.06 & 65.09 & 85.39  & 94.48 & 53.37  & 57.35 & 79.67 & 81.47 & 65.86 & 65.40  & \multicolumn{1}{c|}{20.27} & 56.42  & 24.68   & 34.82   \\ \hdashline
\multicolumn{1}{l|}{BUS ~\cite{choe2024open}}      & \textbf{98.31} & \textbf{74.34}  & \textbf{73.65}  & 48.05 & \textbf{34.62} & 46.21 & \textbf{30.15} & \textbf{74.17} & 87.06  & 95.77 & 64.38  & \textbf{66.91} & \textbf{89.31} & \textbf{87.84} & \textbf{89.77} & \textbf{71.89}  & \multicolumn{1}{c|}{16.25} & \textbf{65.47}  & 29.70   & 41.26   \\ \hline
\multicolumn{1}{l|}{UniMAP \textbf{(Ours)}}     & 98.13 & 62.50  & 76.12  & \textbf{85.74} & 27.48 & \textbf{46.56} & 26.07 & 59.63 & \textbf{90.44}  & \textbf{96.31} & \textbf{65.87}  & 66.85 & 82.83 & 87.08 & 68.33 & 70.27  & \multicolumn{1}{c|}{\textbf{35.45}} & 64.08  & \textbf{34.78}   & \textbf{45.51}   \\
\specialrule{1.5pt}{0pt}{0pt}
\end{tabular}%
}
\vspace{-3mm}
\caption{\label{tab:table-gta_idd} Semantic segmentation performance on GTA5 $\rightarrow$ IDD OPDA-SS benchmarks. Our method outperformed baselines in common, private, and overall performance. White columns show individual common class scores, while ``Common'' in gray columns represents the average performance of common classes. The best results are highlighted in bold.\vspace{-6mm}}
\end{table*}

\section{Experiments}
\label{sec:results}

\subsection{Experimental Setup}

\paragraph{Datasets.}
We evaluated our method on two newly defined OPDA-SS benchmarks: Pascal-Context~\cite{mottaghi2014role} $\rightarrow$ Cityscapes~\cite{cordts2016cityscapes}, and GTA5~\cite{richter2016playing} $\rightarrow$ IDD ~\cite{varma2019idd}, which we introduce to assess universal domain adaptation in more realistic settings involving both source-private and target-private classes.
Pascal-Context $\rightarrow$ Cityscapes is a real-to-real scenario, and Pascal-Context contains both in-door and out-door, while Cityscapes only has a driving scene, so it is a scenario with a considerable amount of source-private classes. We selected 12 classes as common classes and the remaining 7 classes (``pole", ``light", ``sign", ''terrain", ``person", ``rider", and ``train") are treated as target-private classes.
GTA5 $\rightarrow$ IDD is a synthetic-to-real scenario and GTA5 features highly detailed synthetic driving scenes set in urban cityscapes, while IDD captures real-world driving scenarios on diverse roads in India. We used 17 classes as common classes, 2 source-private class (``terrain", ``train"), and 1 target-private class (``auto-rickshaw").

\paragraph{Evaluation Protocols.}
In the OPDA-SS setting, both common class and target-private performance are important, so we evaluate methods using H-Score, which can fully reflect them. The H-score is calculated as the harmonic mean of the common mIoU (mean Intersection-over-Union) and the target-private IoU.

\paragraph{Implementation Details.} 
This method is based on BUS. We used the muli-resolution self-training strategy and training parameter used in MIC~\cite{hoyer2023mic}. The network used a MiT-B5~\cite{xie2021segformer} encoder and was initialized with ImageNet-1k ~\cite{deng2009imagenet} pretrained. The learning rate was 6e-5 for the backbone and 6e-4 for the decoder head, with a weight decay of 0.01 and linear learning rate warm-up over 1.5k steps. EMA factor $\alpha$ was 0.999 and the optimizer was AdamW~\cite{li2020improving}. ImageNet feature Distance ~\cite{hoyer2022daformer}, DACS~\cite{tranheden2021dacs} data augmentation, Masked Image Consistency module ~\cite{hoyer2023mic}, and Dilation-Erosion-based Contrastive Loss~\cite{choe2024open} were used. We also modified some of the BUS methods to suit the OPDA setting. In OpenReMix~\cite{choe2024open}, we applied only Resizing Object except Attaching Private and did not use refinement through MobileSAM~\cite{zhang2023faster}. For rare class sampling~\cite{hoyer2022daformer}, we switched from calculating a distribution based on the existing source and applying it to source sampling to applying it to target sampling based on the target pseudo-label distribution. We trained on a batch of two 512 × 512 random crops for 40k iterations. The hyperparameter are set to: $\tau_p=0.5$, $\tau_t=0.968$, $\lambda_1=0.01$, $\lambda_2=0.01$, $\tau=0.1$, and $T=0.01$.

\begin{figure*}[t] 

\centering
\includegraphics[width=1\textwidth]{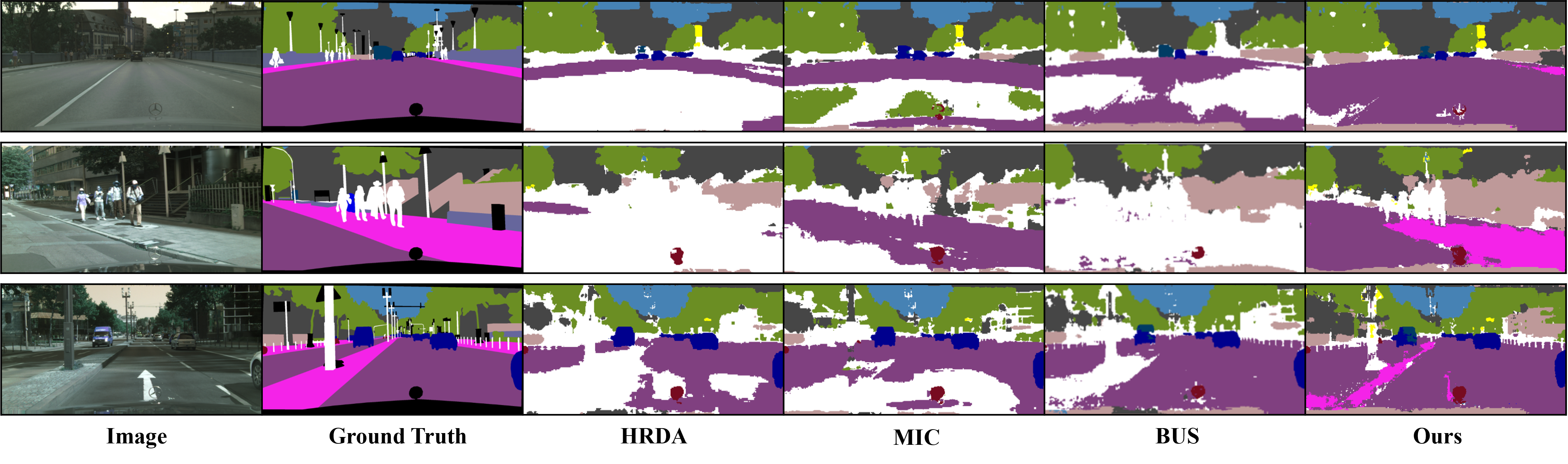}
\vspace{-8mm}

\caption{Qualitative results of OPDA-SS setting. We visualize the segmentation predictions from different methods on the Cityscapes dataset. White and yellow represent target-private and source-private classes, respectively. while other colors indicate common classes (e.g., purple for “road” and pink for “sidewalk”). Compared to HRDA, MIC, and BUS, our method more accurately segments both common and target-private classes.}
\vspace{-6mm}
\label{fig:qual}

\end{figure*} 
\paragraph{Baselines.} 
Since there is no existing research on OPDA-SS, we performed experiments by changing the methods in different settings to suit the OPDA-SS. First, for UniDA for classification methods~\cite{you2019universal, chang2022unified, lu2024mlnet}, we experimented by changing the backbone to a semantic segmentation model. In this case, we used the DeepLabv2~\cite{chen2017deeplab} segmentation network and ResNet-101 ~\cite{he2016deep} as the backbone. For the CDA-SS methods~\cite{hoyer2022daformer,hoyer2022hrda,hoyer2023mic}, we added 1 dimension to the head dimension of the classifier to predict the target-private and assigned an unknown based on the confidence score~\cite{choe2024open}. Lastly, the ODA-SS method, BUS~\cite{choe2024open}, was used as it is.

\subsection{Comparisons with the State-of-the-Art}
We compared performance on two benchmarks for OPDA-SS settings. ~\Cref{tab:table-pascal_city} presents the semantic segmentation performance from Pascal-Context $\rightarrow$ Cityscapes, while ~\Cref{tab:table-gta_idd} presents the performance from GTA5 $\rightarrow$ IDD. As shown in ~\Cref{tab:table-pascal_city}, UniMAP achieved outstanding performance in the Pascal-Context $\rightarrow$ Cityscapes benchmark. Specifically, it outperformed previous approaches by a significant margin, with improvements of approximately 3.3 for Common, 10.89 for Private, and 11.22 in H-score. These results indicate that UniMAP effectively enables the model to learn both common and private classes. Notably, UniMAP surpassed BUS, the state-of-the-art in ODA-SS, in terms of private class performance. Although our method primarily focuses on capturing knowledge of common classes, it also enhances the identification of private classes due to improved representation learning.
In addition, ~\Cref{tab:table-gta_idd} shows the performance comparison for the GTA5 $\rightarrow$ IDD benchmark. Our method demonstrated notable improvements in both Private and H-Score. In particular, while prior methods in CDA-SS showed inferior performance for Private and H-score, our approach led to significant gains of approximately 6.25 for Common, 10.3 for Private, and 9.69 for H-score. Although our method had a relatively lower performance than BUS in Common, it surpassed BUS in Private performance with a margin of about 5.08, ultimately leading to superior H-score results. Overall, the experimental findings demonstrate that our method delivers promising performance in OPDA-SS settings, which is critical for achieving effective UniDA-SS.


\begin{table}[t]
\centering
\resizebox{\columnwidth}{!}{%
\setlength{\tabcolsep}{10.pt}
\renewcommand{\arraystretch}{1.}
\begin{tabular}{cc|cc>{\columncolor[gray]{0.9}}c}
\specialrule{2.pt}{0pt}{0pt}
\multicolumn{2}{c|}{UniMAP} & \multicolumn{3}{c}{Pascal-Context $\rightarrow$ Cityscapes} \\ \hline
DSPD        & TIM        & Common          & Private         & H-Score         \\ \hline
        &       & 53.79           & 26.54           & 36.03           \\ \hdashline
 \checkmark      &       & 59.46           & 27.97           & 38.04           \\
        & \checkmark     & 56.22           & 29.14           & 38.39           \\ \hdashline
 \checkmark      & \checkmark    & \textbf{60.94}           & \textbf{31.27}           & \textbf{41.33}   \\
\specialrule{2.pt}{0.pt}{0pt}
\end{tabular}%
}
\vspace{-3mm}
\caption{\label{tab:ablation1} Ablation study of our method on Pascal-Context $\rightarrow$ Cityscapes. We evaluate the contributions of DSPD and TIM, where the baseline is BUS without private attaching and pseudo-label refinement. The best results are highlighted in bold.}
\vspace{-3mm}
\end{table}

\begin{table}[t]
\centering
\resizebox{\columnwidth}{!}{%
\setlength{\tabcolsep}{10.pt}
\renewcommand{\arraystretch}{1.}
\begin{tabular}{cc|cc>{\columncolor[gray]{0.9}}c}
\specialrule{2.pt}{0pt}{0pt}
\multicolumn{2}{c|}{DSPD} & \multicolumn{3}{c}{Pascal-Context $\rightarrow$ Cityscapes} \\ \hline
$w$        & $\mathcal{L}_{proto}$        & Common          & Private         & H-Score         \\ \hline
&       & 53.79           & 26.54           & 36.03           \\ \hdashline
\checkmark               &         & 54.38           & 21.75           & 31.08           \\ 
              &\checkmark          & \textbf{59.71}           & 26.76           & 36.96           \\ \hdashline
\checkmark              & \checkmark       & 59.46           & \textbf{27.97}           & \textbf{38.04}           \\
\specialrule{2.pt}{0.pt}{0pt}
\end{tabular}%
}
\vspace{-3mm}
\caption{\label{tab:ablation2} Further ablation study of DSPD components on Pascal-Context $\rightarrow$ Cityscapes. $w$ represents pixel-wise weight scaling factor, and $\mathcal{L}_\text{proto}$ represents the prototype loss function. The best results are highlighted in bold.}
\vspace{-3mm}
\end{table}

\begin{table*}[t]
\resizebox{\textwidth}{!}{%
\setlength{\tabcolsep}{5.pt}
\renewcommand{\arraystretch}{1.2}
\begin{tabular}{c|cccccccc|
>{\columncolor[HTML]{EFEFEF}}c |
>{\columncolor[HTML]{EFEFEF}}c }
\specialrule{1.5pt}{0pt}{0pt}
\hline
                         & \multicolumn{8}{c|}{Pascal-Context $\rightarrow$ Cityscapes}                                                                                                                                         & \cellcolor[HTML]{EFEFEF}                                                                                                                        & \cellcolor[HTML]{EFEFEF}                                                                            \\ \cline{2-9}
                         & \multicolumn{3}{c|}{Open Partial Set DA}                              & \multicolumn{3}{c|}{Open Set DA}                                      & \multicolumn{1}{c|}{Partial Set DA} & Closed Set DA  & \cellcolor[HTML]{EFEFEF}                                                                                                                        & \cellcolor[HTML]{EFEFEF}                                                                            \\ \cline{2-9}
\multirow{-3}{*}{Method} & Common         & Private        & \multicolumn{1}{c|}{H-Score}        & Common         & Private        & \multicolumn{1}{c|}{H-Score}        & \multicolumn{1}{c|}{Common}         & Common         & \multirow{-3}{*}{\cellcolor[HTML]{EFEFEF}\begin{tabular}[c]{@{}c@{}}Common\\ Average\end{tabular}} & \multirow{-3}{*}{\cellcolor[HTML]{EFEFEF}\begin{tabular}[c]{@{}c@{}}H-Score\\ Average\end{tabular}} \\ \hline
DAF                      & 54.24          & 4.43           & \multicolumn{1}{c|}{8.19}           & 44.27          & 12.07          & \multicolumn{1}{c|}{18.97}          & \multicolumn{1}{c|}{35.18}          & 46.48          & 44.51                                                                                                                                           & 12.46                                                                                               \\
HRDA                     & 51.89          & 8.55           & \multicolumn{1}{c|}{14.68}          & 52.76          & 14.76          & \multicolumn{1}{c|}{23.07}          & \multicolumn{1}{c|}{51.99}          & 63.17          & 54.76                                                                                                                                           & 18.40                                                                                               \\
MIC                      & 48.67          & 7.85           & \multicolumn{1}{c|}{13.52}          & \textbf{60.88} & 23.79          & \multicolumn{1}{c|}{34.21}          & \multicolumn{1}{c|}{58.04}          & \textbf{65.68} & 57.97                                                                                                                                           & 21.51                                                                                               \\
BUS                      & 57.64          & 20.38          & \multicolumn{1}{c|}{30.11}          & 60.67          & \textbf{27.05} & \multicolumn{1}{c|}{\textbf{37.42}} & \multicolumn{1}{c|}{58.54}          & 60.24          & 59.26                                                                                                                                           & 33.57                                                                                               \\ \hline
{UniMAP \textbf{(Ours)}}            & \textbf{60.94} & \textbf{31.27} & \multicolumn{1}{c|}{\textbf{41.33}} & 58.50          & 24.73          & \multicolumn{1}{c|}{34.76}          & \multicolumn{1}{c|}{\textbf{59.44}} & 64.74          & \textbf{60.86}                                                                                                                                  & \textbf{37.90}                                                                                      \\ \hline

\specialrule{1.5pt}{0pt}{0pt}
\end{tabular}%
}
\caption{\label{tab:ablation3} Experimental results on Pascal-Context $\rightarrow$ Cityscapes for various domain adaptation scenarios. For a fair comparison, all methods used a head-expansion model. The best results are highlighted in bold.}
\vspace{-3mm}
\end{table*}

\subsection{Qualitative Evaluation}
We conducted qualitative experiments under the OPDA-SS setting. ~\Cref{fig:qual} compared prediction maps from Cityscapes against baselines, where white and yellow represent target-private and source-private classes, respectively, while other colors denote common classes. Baseline methods such as HRDA, MIC, and BUS tend to either misclassify common classes as target-private or sacrifice common class accuracy to detect target-private regions. In contrast, UniMAP successfully predicted both common and target-private classes. Notably, it accurately identified the ``sidewalk" class (pink) in rows 2 and 3, unlike other baselines. These results indicate that UniMAP effectively balances the identification of common and target-private classes.

\subsection{Ablation Study}
\paragraph{Ablation Study about UniMAP.}
Table ~\ref{tab:ablation1} shows the experimental results of the ablation study of the performance contribution of each component. As described in the Implementation Details section, the baseline model, derived by removing the Attaching Private and refinement pseudo-label module from the BUS, achieved an H-Score of 36.03. First, applying DSPD alone to the baseline, the H-Score improves to 38.04, increasing both Common and Private performance. This enhancement indicates that DSPD effectively captures domain-specific features, improving performance for both the common and target-private classes compared to the Baseline. Next, when only applying TIM alone to the baseline, also improves performance, achieving an H-Score of 38.39, with better Private. This result suggests that TIM successfully learns domain-invariant representations between source and target by leveraging target pseudo-labels, thus enhancing overall performance. Finally, when both DSPD and TIM are applied to the baseline, the model achieves the best performance, with an H-Score of 41.33. This demonstrates that DSPD and TIM work synergistically, enabling the model to achieve superior performance across both common and target-private classes.

\paragraph{Ablation Study about DSPD.}
Table ~\ref{tab:ablation2} shows the impact of the individual components of DSPD, namely $L_{proto}$ and $w$ on performance in the Pascal-Context $\rightarrow$ Cityscapes scenario.
The $L_{proto}$ represents the pixel embedding loss in the ETF space, designed to guide pixel embeddings within a class to be closer to their respective prototypes. When only $L_{proto}$ is applied, the model achieves a Common of 59.71, a Private of 26.76, and an H-Score of 36.96. This result suggests that $L_{proto}$ alone can enhance the clustering of pixel embeddings around domain-specific prototypes, thereby improving overall performance compared to the baseline.
The $w$, on the other hand, means a weighting mechanism based on the ETF prototype structure that estimates the common class more effectively and applies weights scaling accordingly. When only $w$ is used, the Common 
 drops to 54.38, and the Private score falls to 21.75, resulting in a lower H-Score of 31.08. This indicates that while $w$ is utilized in distinguishing common classes, it is less effective without the guidance provided by $L_{proto}$. When both $L_{proto}$ and $w$ are combined, the model achieves the best performance, with a Common of 59.46, a Private of 27.97, and an H-Score of 38.04. This demonstrates that the two components are complementary: $L_{proto}$ enhances pixel embedding alignment with domain-specific prototypes, while $w$ further boosts the ability to focus on common class pixels with appropriate weighting. Together, they yield a notable improvement in the overall H-Score.

\paragraph{Comparisons in Various Category Settings.}
We further compared the performance and generalization ability of UniMAP across various domain adaptation settings. As shown in ~\ref{tab:ablation3}, while some existing methods achieve slightly better results in Closed Set and Open Set settings due to their specialized assumptions, UniMAP demonstrates clear advantages in Partial Set and Open Partial Set, where prior methods have not been actively explored. Notably, UniMAP achieves the highest scores, with a Common Average of 60.86 and an H-Score Average of 37.90, validating its robustness and effectiveness across varying category shift configurations. These results highlight the practicality of our framework for the real-world scenario, where category settings are often unknown.

\section{Conclusion}
\label{sec:conclusions}

In this paper, we proposed a new framework for UniDA-SS, called UniMAP. Since UniDA-SS must handle different domain configurations without prior knowledge of category settings, it is very important to identify and learn common classes across domains. To this end, UniMAP incorporates two key components: Domain-Specific Prototype-based Distinction (DSPD) and Target-based Image Matching (TIM). DSPD is used to estimate common classes from the unlabeled target domain, while TIM samples labeled source images to transfer knowledge to the target domain effectively. Experimental results show that our method improved average performance across different domain adaptation scenarios. We hope our approach sheds light on the necessity of universal domain adaptation for the semantic segmentation task.

\section*{Acknowledgment}
\label{sec:acknowledgment}
This research was conducted with the support of the HANCOM InSpace Co., Ltd. (Hancom-Kyung Hee Artificial Intelligence Research Institute), and was supported by Korea Planning \& Evaluation Institute of Industrial Technology (KEIT) grant funded by the Korea government (MOTIE) (RS-2024-00444344), and in part by Institute of Information \& communications Technology Planning \& Evaluation (IITP) grant funded by the Korea government (MSIT) (No. RS-2019-II190079, Artificial Intelligence Graduate School Program (Korea University)).
{
    \small
    \bibliographystyle{ieeenat_fullname}
    \bibliography{main}
}

\end{document}